\documentclass{article}
\usepackage{amsmath,epsfig}
\usepackage{amssymb}
\usepackage{subfigure}
\usepackage{algorithm}
\usepackage{algorithmicx}
\usepackage{algpseudocode}
\usepackage{setspace}
\usepackage{longtable}
\usepackage{diagbox}
\usepackage{multirow}
\usepackage{color}
\usepackage[switch]{lineno}
\usepackage[preprint]{spconfa4}

\copyrightnotice{978-1-6654-3864-3/21/\$31.00 ©2021 IEEE}

\let\OLDthebibliography\thebibliography
\renewcommand\thebibliography[1]{
  \OLDthebibliography{#1}
  \setlength{\parskip}{0pt}
  \setlength{\itemsep}{0pt plus 0.3ex}
}

\begin{document}\sloppy

\def\x{{\mathbf x}}
\def\L{{\cal L}}

\title{Transferable Adversarial Examples for Anchor Free Object Detection}
%
\name{
Quanyu Liao$^1$, Xin Wang$^2{}^{\dagger}$\thanks{{\bf $\dagger$} Corresponding authors: Xin Wang (xinw@keyamedna.com), Xi Wu (xi.wu@cuit.edu.cn).
This work was supported by Sichuan Science and Technology Program 2019ZDZX0007, 2019YFG0399, and 2019YFG0496.}, Bin Kong$^2$,  Siwei Lyu$^3$, \textit{Bin Zhu}$^4$, \textit{Youbing Yin}$^2$, \textit{Qi Song}$^2$, \textit{Xi Wu}$^1{}^{\dagger}$
}
\address{
$^1$ Chengdu University of Information Technology, Chengdu, China \\
$^2$ Keya Medical, Seattle, USA\\
$^3$ University at Buffalo, State University of New York, USA \\
$^4$ Microsoft Research Asia, Beijing, China 
}

\maketitle

\begin{abstract}
Deep neural networks have been demonstrated to be vulnerable to adversarial attacks: subtle perturbation can completely change prediction result. The vulnerability has led to a surge of research in this direction, including adversarial attacks on object detection networks. However, previous studies are dedicated to attacking anchor-based object detectors. In this paper, we present the first adversarial attack on anchor-free object detectors. It conducts category-wise, instead of previously instance-wise, attacks on object detectors, and leverages high-level semantic information to efficiently generate transferable adversarial examples, which can also be transferred to attack other object detectors, even anchor-based detectors such as Faster R-CNN. Experimental results on two benchmark datasets demonstrate that our proposed method achieves state-of-the-art performance and transferability.
\end{abstract}
\begin{keywords}
Category-wise attacks, adversarial attacks, object detection, anchor-free object detection
\end{keywords}
\section{Introduction}
The development of deep neural network has significantly improved the performance of many computer vision tasks. However, many recent works show that deep-learning-based algorithms are vulnerable to adversarial attacks~\cite{carlini2017towards,dong2018boosting,xie2019improving,croce2019minimally,dong2019evading}. The vulnerability of deep networks is observed in many different problems~\cite{bose2018adversarial,chen2018robust}, including object detection, one of the most fundamental tasks in computer vision.

Regarding the investigation of the vulnerability of deep models in object detection, previous efforts mainly focus on classical anchor-based networks such as Faster-RCNN~\cite{ren2015faster}. However, the performance of these anchor-based networks is limited by the choice of anchor boxes. Fewer anchors lead to faster speed but lower accuracy. Thus, advanced anchor-free models such as CornerNet~\cite{law2019cornernet} and CenterNet~\cite{zhou2019objects} are becoming increasingly popular, achieving competitive accuracy with traditional anchor-based models yet with faster speed and stronger adaptability. However, to the best of our knowledge, there is no published work on investigating the vulnerability of anchor-free networks.

\begin{figure}[t]
\begin{center}
\includegraphics[width=2.9in]{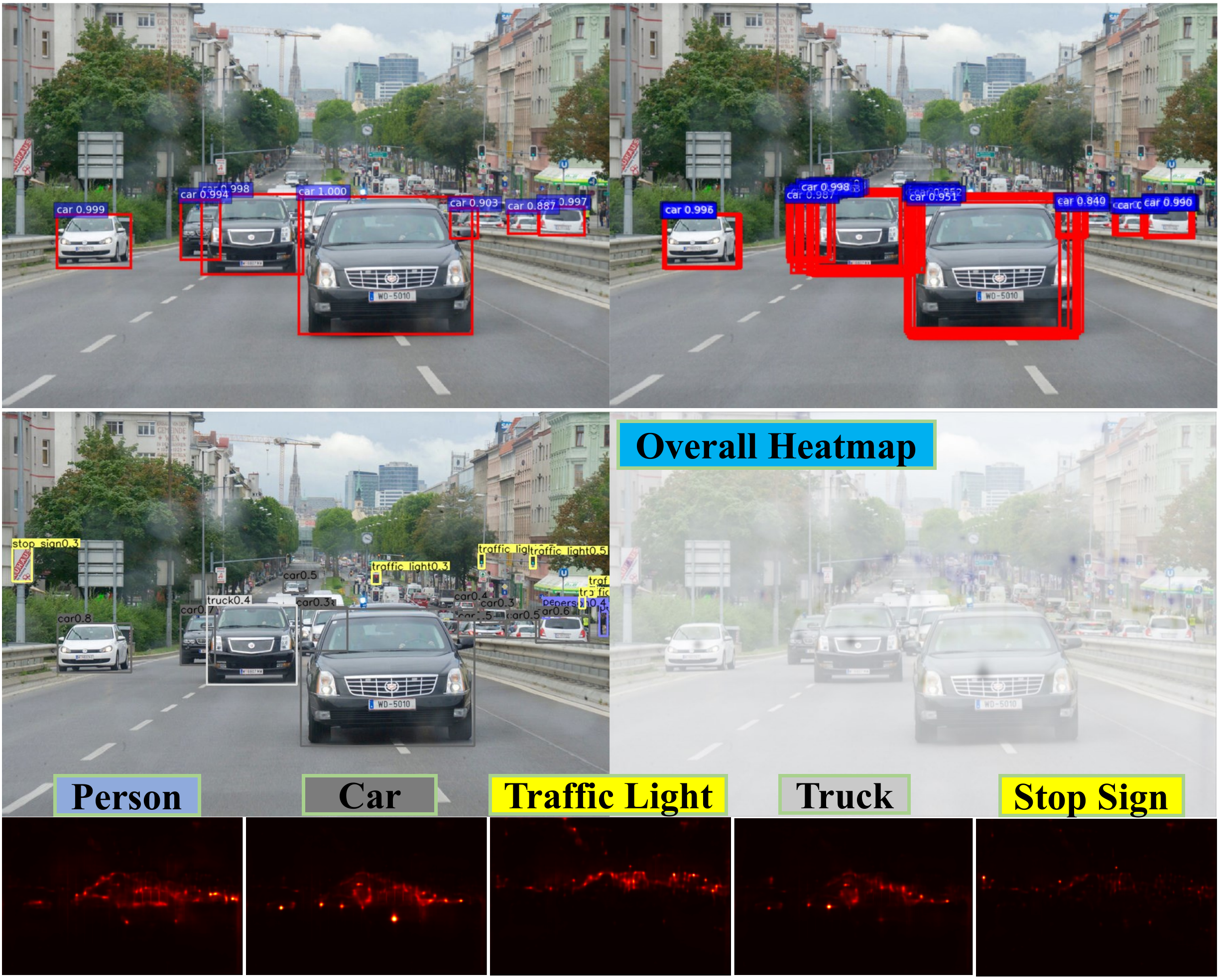}
\end{center}
\vspace*{-0.3cm}
\caption{ \small
\textbf{First row:} The detected results (left) and the proposals (right) of Faster R-CNN~\cite{ren2015faster}.
\textbf{Second row:} The detected results (left) and the overall heatmap (right) of CenterNet~\cite{zhou2019objects}. \textbf{Third row:} Selected target pixels (red) for each category by our method.}
\label{fig_pix_select}
\vspace*{-0.5cm}
\end{figure}

Previous work DAG~\cite{xie2017adversarial} achieved high white-box attack performance on the FasterRCNN, but DAG is hardly to complete an effective black-box attack. DAG also has the disadvantages of high time-consuming, these two shortcomings make DAG difficult to be used in real scenes. These two shortcomings of DAG principally because DAG only attacks one proposal in each attack iteration. It will make the generated adversarial perturbation only effective for one proposal, which leads to bad transferring attack performance and consumes an amount of iterations to attack all objects.

Meanwhile, attack an anchor-based detector is unlike to attack an anchor-free detector, which select top proposals from a set of anchors for the objects, anchor-free object detectors detect objects by finding objects' keypoints via the heatmap mechanism (see Fig.~\ref{fig_pix_select}), using them to generate corresponding bounding boxes, and selecting the most probable keypoints to generate final detection results. This process is completely different from anchor-based detectors, making anchor-based adversarial attacks unable to directly attack anchor-free detectors.

To solve above two problems, we propose a novel algorithm, \emph{Category-wise Attack (CW-Attack)}, to attack anchor-free object detectors. It attacks all instances in a category simultaneously by attacking a set of target pixels in an image, as shown in Fig.~\ref{fig_pix_select}.
The target pixel set includes not only all detected pixels, which are highly informative pixels as they contain higher-level semantic information of the objects, but also ``runner-up pixels" that have a high probability to become rightly detected pixels under small perturbation.
Our approach guarantees success of adversarial attacks.
Our CW-Attack is formulated as a general framework that minimizes $L_p$ of perturbation, where $L_p$ can be $L_0$, $L_1$, $L_2$, $L_\infty$, etc., to flexibly generate different types of perturbations, such as dense or sparse perturbations.
Our experimental results on two benchmark datasets, PascalVOC \cite{everingham2015pascal} and MS-COCO \cite{lin2014microsoft}, show that our method outperforms previous state-of-the-art methods and generates robust adversarial examples with superior transferability.

Our CW-Attack disables object detection by driving feature pixels of objects into wrong categories. This behavior is similar to but the essence is completely different from attacking semantic segmentation approaches~\cite{xie2017adversarial}. First, they have different targets to optimize: the goal is to change the category of an object's bounding box in our attack and a detected pixel's category in attacking semantic segmentation. Second, they have different relationships to attack success: once pixels have changed their categories, the attack is successful for attacking semantic segmentation but not yet for our attack. As we will see in Fig.~\ref{select_pixel}, objects can still be detected even when all heatmap pixels have been driven into wrong categories.

This paper has the following major contributions:
\textbf{(i)} We propose the first adversarial attack on anchor-free object detection. It attacks all objects in a category simultaneously instead of only one object at a time, which avoids perturbation over-fitting on one object and increases transferability of generated perturbation.
\textbf{(ii)} Our CW-Attack is designed as a general $L_p$ norm optimization framework. When minimizing perturbation's $L_0$ norm (see Sec.~\ref{sec: sca}), it generates sparse adversarial samples by only modifying less than 1\% pixels. While minimizing its $L_\infty$ norm (detail in supplement materials), it can attack all objects of \emph{all} categories simultaneously, which further improves the attacking efficiency.
\textbf{(iii)} Our method generates more transferable and robust adversarial examples than previous attacks. It achieves the state-of-the-art attack performance for both white-box and black-box attacks on two public benchmark datasets, MS-COCO and PascalVOC.

\section{Our Category-wise Attack}
\label{sec: approach}

In this section, we first define the optimization problem of attacking anchor-free detectors and then provide a detailed description of our Category-wise Attack (CW-Attack).

\label{Problem_Formulation} 
\noindent \textbf{Problem Formulation.} Suppose there exist $k$ object categories, $\{C_1, C_2, ..., C_k\}$, with detected object instances. We use $S_{target}$ to denote the target pixel set of category $C_{target}$ whose detected object instances will be attacked, leading to $k$ target pixel sets: $\{S_1, S_2, ..., S_k\}$. The category-wise attack for anchor-free detectors is formulated as the following constrained optimization problem:
\begin{equation}
   \begin{aligned}
       \mathop{\text{minimize}} \limits_{r} \quad & \Vert{r} \Vert_{p} \\
       s.t. ~ \quad & ~\forall k, s\in S_{target}\in \{S_1, S_2, ..., S_k\}\\
       &{\arg\max}_n \{f_n(x+r, s)\} \not = C_{target}
   \end{aligned}
   \label{categorywiseoptimization}
\end{equation}
where $r$ is an adversarial perturbation, $\Vert{\cdot}\Vert_{p}$ is the $L_p$ norm, $p \in \{0, 1, 2,\infty\}$, $x$ is a clean input image, $x+r$ is an adversarial example, $f(x+r, s)$ is the classification score vector~(logistic) and $f_n(x+r, s)$ is its $n^{th}$ value, 
${\arg\max}_n\{f_n(x+r, s)\})$ denotes the predicted object category on a target
pixel $s \in S_{target}$ of adversarial example $x+r$.

\begin{figure}[t]
  \begin{center}
  \includegraphics[width=0.9\linewidth]{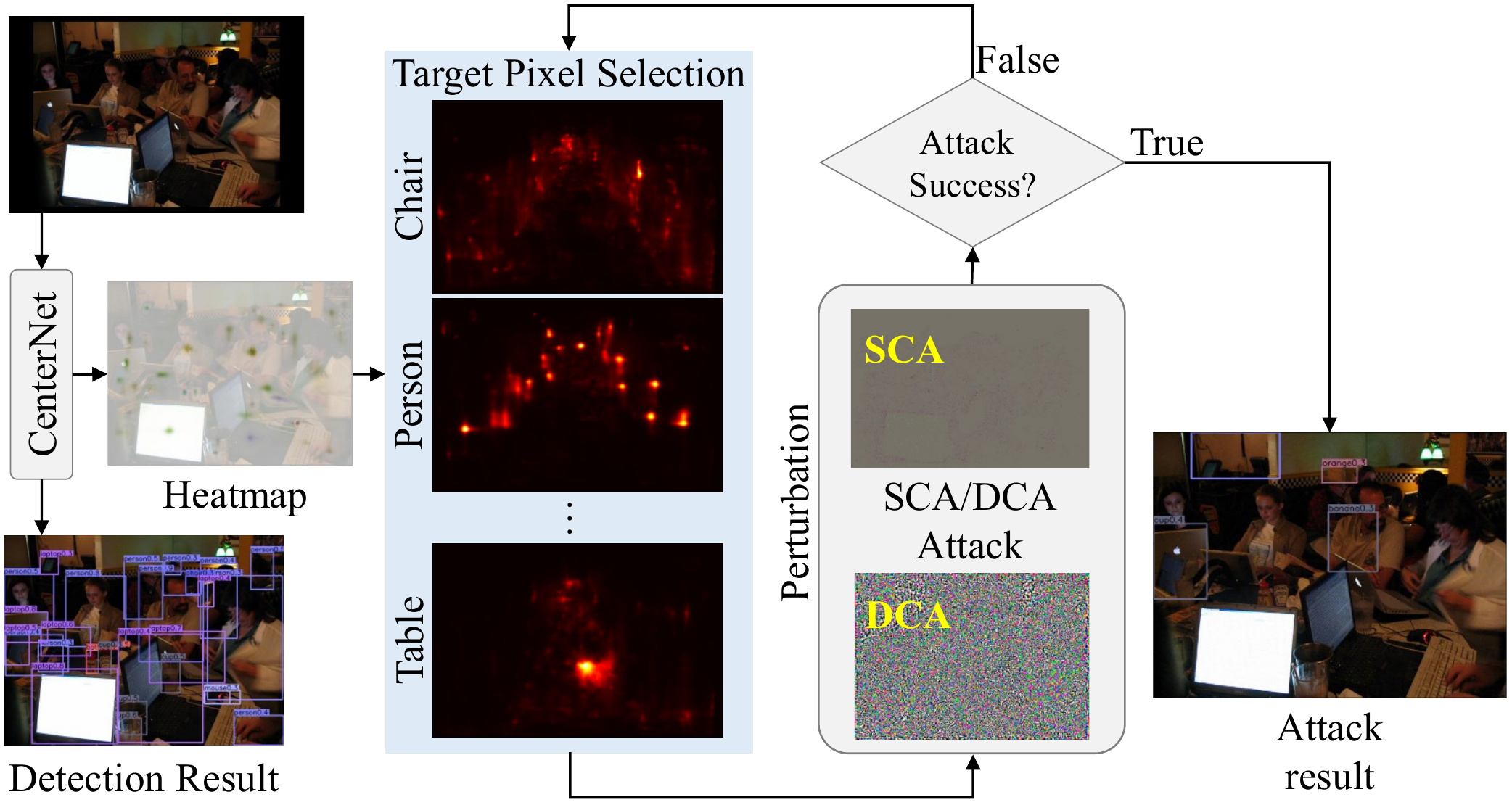}
  \end{center}
  \vspace*{-0.3cm}
  \caption{ \small
  Overview of CW-Attack. Target pixel sets $\{S_1, S_2, ..., S_k\}$ are first extracted from the heatmap for all object categories. SCA or DCA is then used to generate perturbation, depending on minimizing perturbation's $L_0$ or $L_\infty$ norm. Finally, we check whether the attack is successful. If not, a new perturbation is generated from the current adversarial example in the next iteration. }
  \label{fig_overview_illustration}
  \vspace*{-0.5cm}
\end{figure}

The overview of the proposed CW-Attack is shown in Fig.~\ref{fig_overview_illustration}. In the following description of our method, we assume the task is a non-target multi-class attack. If the task is a target attack, our method can be described in a similar manner.

\noindent \textbf{Category-wise Target Pixel Set Selection.}
\label{Pixels_Selection} In solving our optimization problem~(\ref{categorywiseoptimization}), it is natural to use all \emph{detected pixels} of category $C_{target}$ as target pixel set $S_{target}$.  The {detected pixels} are selected from the heatmap of category $C_{target}$ generated by an anchor-free detector such as CenterNet \cite{zhou2019objects} with their probability scores higher than the detector's preset visual threshold and being detected as right objects. Unfortunately, it does not work. After attacking all detected pixels into wrong categories, we expect that the detector should not detect any correct object, yet our experiments with CenterNet turn out that it still can. 

\begin{figure}[t]
    \begin{center}
    \includegraphics[width=3.1in]{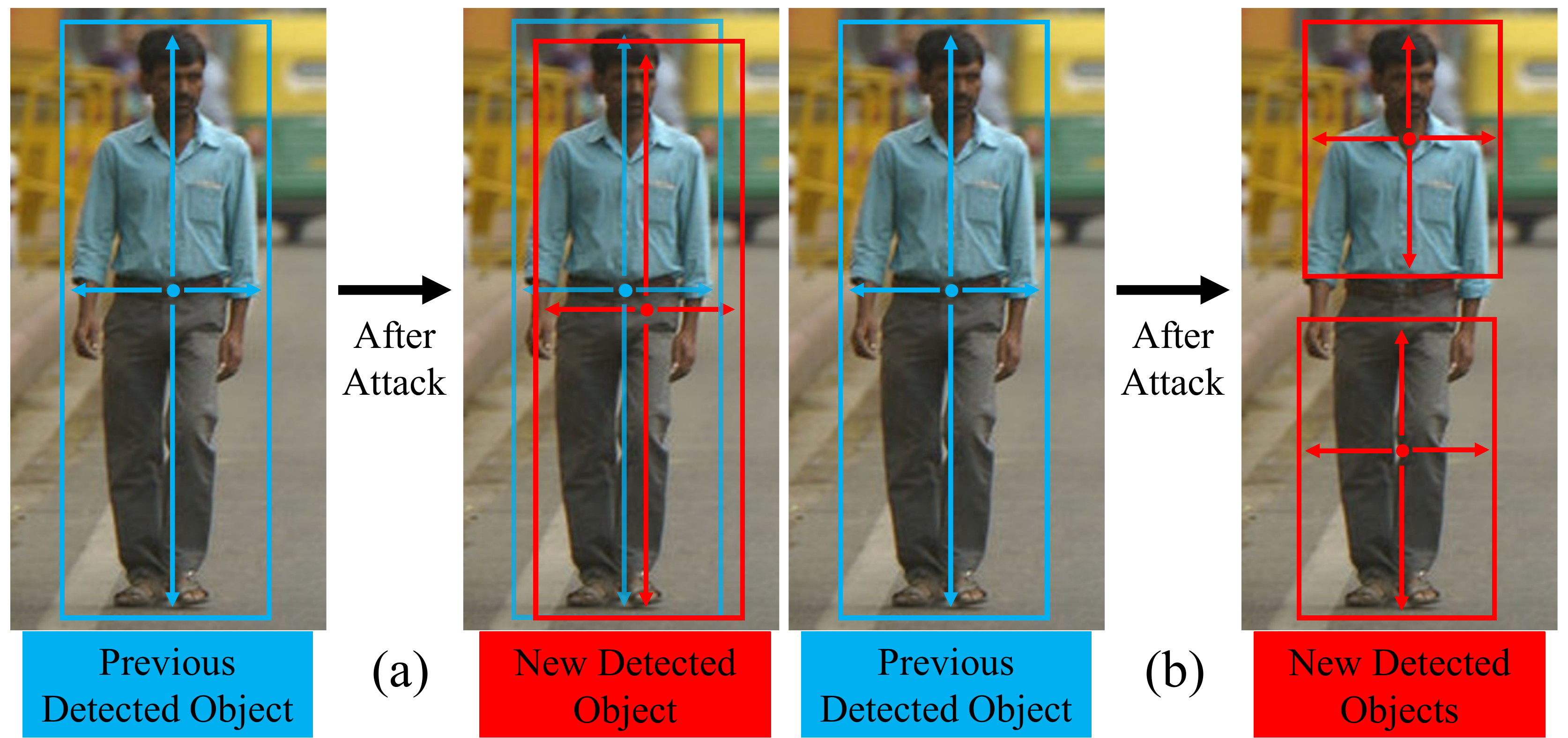}
    \end{center}
    \vspace*{-0.5cm}
    \caption{ \small
    \textbf{Blue} points denote originally detected keypoints before the attack.
    \textbf{Red} points denote newly detected keypoint after the attack.
    \textbf{(a)-Left \& (b)-Left:} a detected object and a detected keypoint at the center of the person before the attack.
    \textbf{(a)-Right \& (b)-Right:} detection results after attacking only detected pixels. After attacking all detected pixels, a neighboring pixel of the previously detected keypoint is detected as the correct object for \textbf{(a)-Right}, and the centers of the top half and the bottom half of the person appear as newly detected keypoints still detected as a person for \textbf{(b)-Right}. In both cases, mAP is barely reduced.}
    \label{select_pixel}
    \vspace*{-0.5cm}
\end{figure}

Further investigation reveals two explanations: \textbf{(1)} Neighboring background pixels of the heatmap not attacked can become detected pixels with the correct category. Since their detected box is close to the old detected object, CenterNet can still detect the object even though all the previously detected pixels are detected into wrong categories. An example is shown in Fig.~\ref{select_pixel}-(a).
{\textbf{(2)} CenterNet regards center pixels of an object as keypoints. After attacking detected pixels located around the center of an object, newly detected pixels may appear in other positions of the object, making the detector still be able to detect multiple local parts of the correct object with barely reduced mAP.
An example is shown in Fig.~\ref{select_pixel}-(b).}

Pixels that can produce one of the above two changes are referred to as \emph{runner-up pixels}. We find that almost all \emph{runner-up pixels} have a common characteristic: their probability scores are only a little below the visual threshold. Based on this characteristic, our CW-Attack sets an attacking threshold, $t_{attack}$, lower than the visual threshold, and then selects all the pixels from the heatmap whose probability score is above $t_{attack}$ into $S_k$. This makes $S_k$ include all detected pixels and \emph{runner-up pixels}. Perturbation generated in this way can also improve robustness and transferable attacking performance.

\begin{figure}[t]
   \begin{center}
   \includegraphics[width=.7\linewidth]{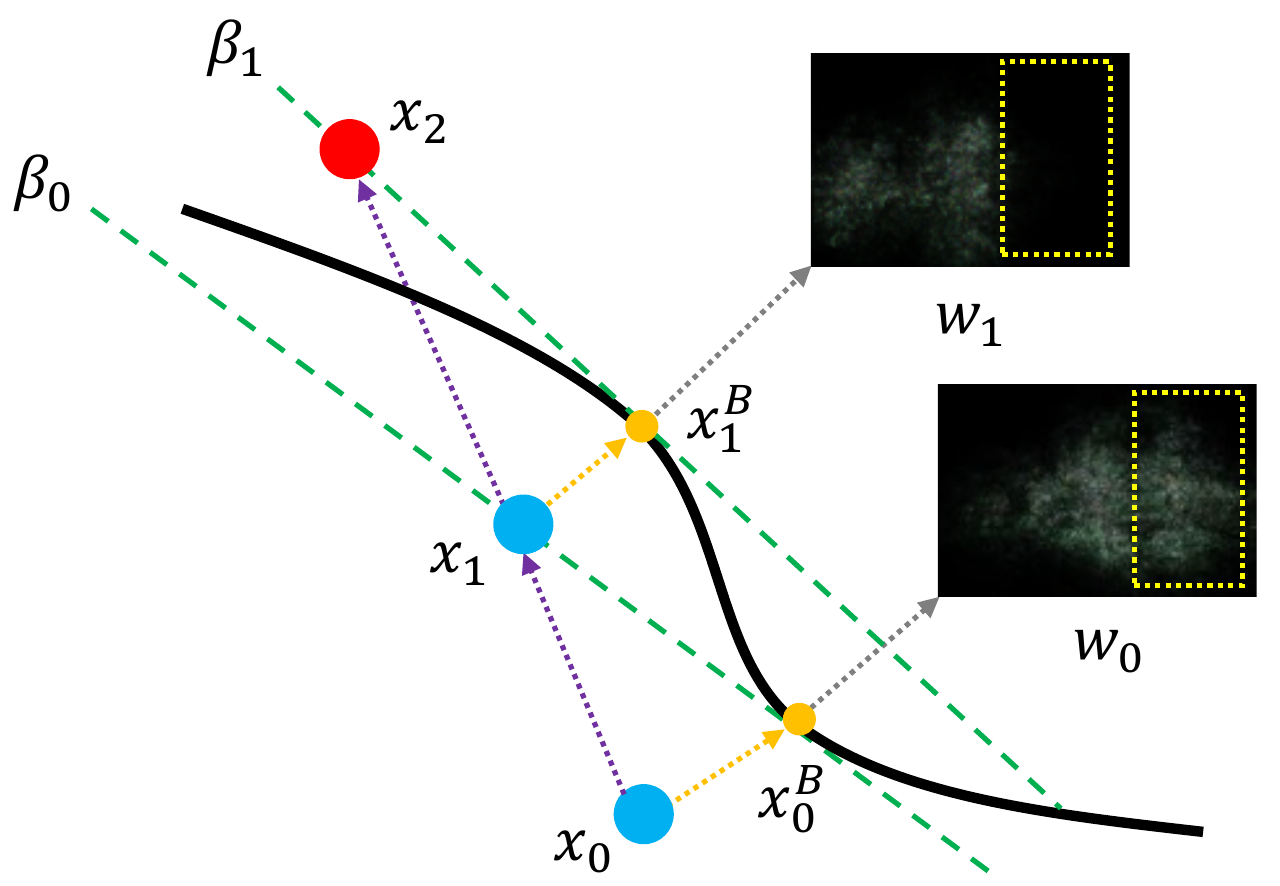}
   \end{center}
   \vspace*{-0.3cm}
   \caption{ \small
   Illustration of SCA with the `Car' category of Fig.~\ref{fig_pix_select}. The black solid line denotes the real decision boundary of the object detector. Blue points denote adversarial examples that have not attacked all objects successfully. Red point denote adversarial example that have already attacked all objects successfully. This figure illustrates two iterations of the attack, $x_0 \rightarrow x_1$ and $x_1 \rightarrow x_2$. Take $x_0 \rightarrow x_1$ for example, SCA first generates dense adversarial example $x_0^B$ (yellow point) by \textit{CW-DF} and approximated linear decision boundary $\beta_0$ (green dash lines). Then it uses \textit{LinearSolver} (purple dash lines) to add a sparse perturbation to support $x_0$ to approximate decision boundary $\beta_0$ by satisfying $\beta \ \mathop{=} \ \{x:w^T(x - x^B) = 0 \}$ until a valid sparse adversarial example $x_1$ is obtained.
   The two images are the visualization of the normal vector $\mathop{w}$, and the yellow boxes on the two images indicate that the weights for the 'Car' object are reduced. }
   \vspace*{-.5cm}
   \label{fig_sca_car}
\end{figure}

\label{sec: sca}
\noindent \textbf{Sparse Category-wise Attack.} The goal of the sparse attack is to fool the detector while perturbing a minimum number of pixels in the input image. It is equivalent to setting $p=0$ in our optimization problem (\ref{categorywiseoptimization}), i.e. minimizing $\Vert{r}\Vert{}_0$ according to $S_{target}$. Unfortunately, this is an NP-hard problem. To solve this problem,  SparseFool~\cite{modas2019sparsefool} relaxes this NP-hard problem by iteratively approximating the classifier as a local linear function in generating sparse adversarial perturbation for image classification. 

Motivated by the success of SparseFool on image classification, we propose Sparse Category-wise Attack (SCA) to generate sparse perturbations for anchor-free object detectors. It is an iterative process. In each iteration, one target pixel set is selected from category-wise target pixel sets to attack.

More specifically, given an input image $x$ and current category-wise target pixel sets $\{S_1, S_2, ..., S_k\}$, SCA selects the pixel set that has the highest probability score from $\{S_1, S_2, ..., S_k\}$ as target pixel set $S_{target}$ and use \textit{Category-Wise DeepFool (CW-DF)}\footnote{See the supplement materials for the detail of \textit{CW-DF}, \textit{ApproxBoundary}, \textit{LinearSolver} and \textit{RemovePixels}.} to generate dense adversarial example $x^B$ by computing perturbation on $S_{target}$. 
CW-DF is adapted from DeepFool~\cite{moosavi2016deepfool} to become a category-wise attack algorithm for anchor-free object detection.

Then, SCA uses the \textit{ApproxBoundary} to approximate the decision boundary, which is locally approximated with a hyperplane $\beta$ passing through $x^B$:

\vspace{-0.3cm}
\begin{equation}
   \begin{aligned}
       & \beta \ \mathop{=}^{\triangle} \ \{x:w^T(x - x^B) = 0 \},
    \end{aligned}
\end{equation}
where $w$ is the normal vector of hyperplane $\beta$ and approximated with the following equation~\cite{modas2019sparsefool}:
\begin{equation}
   \begin{aligned}
       w :=~\nabla\sum_{i=1}^n &  {f_{\text{argmax}_n{f_n(x^B, s)}}(x^B, s)} \\
       &- \nabla\sum_{i=1}^n{f_{\text{argmax}_n{f_n(x, s)}}(x^B, s)}.
\end{aligned}
\end{equation}

{The sparse adversarial perturbation} can then be computed via the \textit{LinearSolver} process~\cite{modas2019sparsefool}. The process of generating perturbation through the \textit{ApproxBoundary} and the \textit{LinearSolver} of SCA is illustrated in Fig.~\ref{fig_sca_car}.

After attacking $S_{target}$, SCA uses \textit{RemovePixels} to update $S_{target}$ by removing the pixels that are no longer detected. Specifically, it takes $x_{i,j}$, $x_{i,j+1}$, and $S_{target}$ as input. \textit{RemovePixels} first generates a new heatmap for perturbed image $x_{i,j+1}$ with the detector. Then, it checks whether the probability score of each pixel in $S_{target}$ is still higher than $t_{attack}$ on the new heatmap. Pixels whose probability score is lower than $t_{attack}$ are removed from $S_{target}$, while the remaining pixels are retained in $S_{target}$. Target pixel set $S_{target}$ is thus updated. If $\{S_1, S_2, ..., S_k\} \in \varnothing$, which indicates that no {correct object}
can be detected after the attack, the attack for all objects of $x$ is successful, and we output the generated adversarial example.

\renewcommand{\algorithmicrequire}{\textbf{Input:}}
\renewcommand{\algorithmicensure}{\textbf{Output:}}
\begin{algorithm}[t]
    \setstretch{1.3}
    \caption{Sparse Category-wise Attack (SCA)}
    \label{sca}
    \begin{algorithmic}

        \Require
        image $x$, target pixel set $\{S_1, S_2, ..., S_k\}$, \\
        available categories $\{C_1, C_2, ..., C_k\}$
        \Ensure
        perturbation $r$

        \State{Initialize: $x_1 \leftarrow x, i \leftarrow 1, j \leftarrow 1, S_0 \leftarrow S$}

        \While{$\{S_1, S_2, ..., S_k\} \not\in \varnothing$}
        
            \State$target = \text{argmax}_k\sum_{s\in S_k} \text{softmax}_{C_k}~f(x_i, s)$
            
            \State$S_{target, 1} \leftarrow S_{target}$

            \State$x_{i,j} \leftarrow x_i$

            \While{$j \leq M_s$ or $S_{target, j} \in \varnothing $}

                \State$x_j^B = \text{CW-DF}~(x_{i,j})$
                \State$w_j = \text{ApproxBoundary}~(x_j^B, S_{target, j})$
                \State$x_{i, j+1} = \text{LinearSolver}~(x_{i,j},w_j, x_j^B)$
                \State$S_{target} = \text{RemovePixels}~(x_{i,j}, x_{i,j+1}, S_{target})$

                \State$j = j + 1$

            \EndWhile

            \State$x_{i+1} \leftarrow x_{i,j}$
            \State$i = i + 1$

        \EndWhile
 
        \State{\textbf{return} $r = x_i - x_1$ }
        
    \end{algorithmic}
    \label{algorithm_sca}
\end{algorithm}

The SCA algorithm is summarized in Alg.~\ref{algorithm_sca}. Note that SCA will not fall into an endless loop. In an iteration, if SCA fails to attack any pixels of $S_{target}$ in the inner loop, SCA will attack the same $S_{target}$ in the next iteration. During this process, SCA keeps accumulating perturbations on these pixels, with the probability score of each pixel in $S_{target}$ keeping reducing, until the probability score of every pixel in $S_{target}$ is lower than $t_{attack}$. By then, $S_{target}$ is attacked successfully.

\label{sec: dca}
\noindent \textbf{Dense Category-wise Attack.} It is interesting to investigate our optimization problem (\ref{categorywiseoptimization}) for $p=\infty$. FGSM~\cite{goodfellow2014explaining} and PGD~\cite{madry2017towards} are two most widely used attacks by minimizing $L_{\infty}$. PGD iteratively takes smaller steps in the direction of the gradient. It achieves a higher attack performance and generates smaller $L_{\infty}$ perturbations than FGSM. Our adversarial perturbation generation procedure is base on PGD and is named as \emph{Dense Category-wise Attack} (DCA) since it generates dense perturbations compared to SCA. 

Given an input image $x$ and category-wise target pixel sets $\{S_1, S_2, ..., S_k\}$, DCA\footnote{DCA is summarized in Alg.~5 in the supplement materials. Fig.~1 in the supplement materials shows the perturbation generation process of DCA.} applies two iterative loops to generate adversarial perturbations: each inner loop iteration $j$ computes the local gradient for each category $S_j$ and generates a total gradient for all detected categories; while each outer loop iteration $i$ uses the total gradient generated in the inner loop iteration to generate a perturbation for all the objects of all detected categories.

Specifically, in each inner loop iteration $j$, DCA computes the gradient for every pixel in $S_j$ to attack all object instances in $C_j$ as follows:
DCA first computes the total loss of all pixels in target pixel set $S_j$ corresponding to each available category $C_j$:
\begin{equation}
    \begin{aligned}
        \textit{loss}_{\textit{sum}} = \sum_{s\in S_{\text{j}}}\text{CrossEntropy}~(f(x_i, s),~C_j),
    \end{aligned}
\end{equation}
and then computes local adversarial gradient $r_j$ of $S_j$ on $loss_{sum}$ and normalizes it with $L_\infty$, yielding $r_j'$:
\begin{equation}
   \begin{aligned}
       r_j = & \bigtriangledown_{x_i}\textit{loss}_{\textit{sum}},~~~
       r_j' = & \frac{r_j}{\Vert r_j\Vert_{\infty}}.
\end{aligned}
\end{equation}

\begin{figure*}[t]
    \centering
    \includegraphics[width=1.0\linewidth]{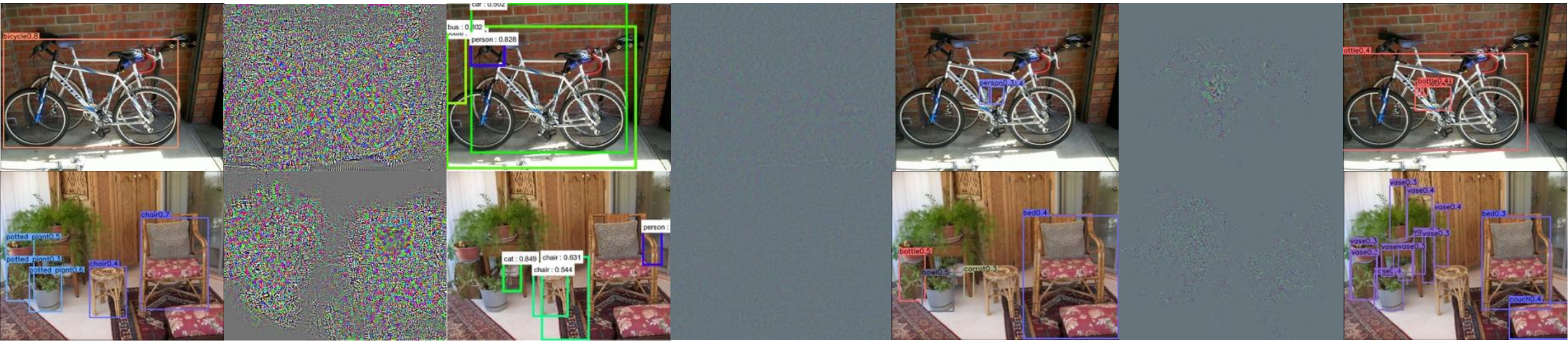}
    \vspace{-0.4cm}
    \caption{\small
    Qualitative comparison between DAG and our methods. Each row is an example. \textbf{Column 1:} Detection results of clean inputs on CenterNet. \textbf{Column 2$\&$3:} DAG perturbations and DAG attacked results on Faster-RCNN. \textbf{Column 4$\&$5:} DCA perturbations and DCA attacked results on CenterNet. \textbf{Column 6$\&$7:} SCA perturbations and SCA attacked results on CenterNet. Note that in \textbf{Column 6}, the percentage of perturbed pixels for the SCA perturbations is 3.4\% and 3.51\% from top to bottom. We can see that the perturbations of DCA and SCA are smaller than  DAG's. Notably, the proposed SCA changes only a few percentage of pixels. To better show perturbations, we have multiplied the intensity of all perturbation images by 10.}
\label{fig: effi_framework}
\vspace{-0.75cm}
\end{figure*}

\noindent After that, DCA adds up all $r_j'$ to generate total adversarial gradient $G$. Finally, in the outer loop iteration $i$, DCA computes perturbation $pert_i$ by applying $sign$ operation to the total adversarial gradient $G$ \cite{madry2017towards}:
\begin{equation}
   \begin{aligned}
       pert_i = \frac{\epsilon_D}{M_D}\cdot sign(G),
    \end{aligned}
\end{equation}
where $M_D$ denotes the maximum number of cycles of the outer loop, term $\frac{\epsilon_D}{M_D}$ is optimal max-norm constrained weight to constraint the amplitude of $pert_i$ \cite{goodfellow2014explaining}. At the end of the outer loop, DCA uses $RemovePixels$~ to remove the target pixels that have already been attacked successfully on $x_{i+1}$ from of $\{S_1, S_2, ..., S_k\}$.

Since an adversarial perturbation in DCA is generated from normalized adversarial gradients of all categories' objects, DCA attacks all object instances of all the categories simultaneously. It is more efficient than SCA.

\section{Experimental Evaluation}
\label{sec: experiment}

\noindent \textbf{Dataset.} Our method is evaluated on two object detection benchmarks: PascalVOC \cite{everingham2015pascal} and MS-COCO~\cite{lin2014microsoft}.

\noindent \textbf{Evaluation Metrics.} \textbf{i)} \textit{Attack Success Rate (ASR)}: $ASR = 1-mAP_{attack} / {mAP_{clean}}$, where $mAP_{attack}$ and $mAP_{clean}$ are the $mAP$ of the adversarial example and the clean input, respectively.
\textbf{ii)} \textit{Attack Transfer Ratio (ATR)}: It is evaluated as follows: $ATR = ASR_{target}/{ASR_{origin}}$, where $\mathop{ASR_{target}}$ is the $\mathop{ASR}$ of the target object detector to be black-box attacked, and $\mathop{ASR_{origin}}$ is the $\mathop{ASR}$ of the detector that generates the adversarial example.
\textbf{iii)} \textit{Perceptibility}: The perceptibility of an adversarial perturbation is quantified by its $P_{L_2}$ and $P_{L_0}$ norm. \textbf{a) $P_{L_2}$:} $P_{L_2} = \sqrt{{1/N}\sum{r_N^2}}$, where the $\mathop{N}$ is the number of image pixels. We normalize the $\mathop{P_{L_2}}$ from $\mathop{[0,255]}$ to $\mathop{[0,1]}$. \textbf{b) $P_{L_0}$:} $P_{L_0}$ is computed by measuring the proportion of perturbed pixels after attack.

\begin{table}[t]
\centering
    \caption{\small White-box performance comparison. The top row denotes the metrics. Clean and Attack denote the mAP of clean input and adversarial examples, respectively. Time is the average time to generate an adversarial example.}
    \label{overall_white}
    \resizebox{0.8\columnwidth}{!}{
        \begin{tabular}{l|l||l|c|c|c|c}
        \hline
                        & Method& Network         & Clean       & Attack     & ASR       & Time (s)             \\ \hline \hline
        \multirow{6}{*}{\rotatebox[origin=c]{90}{PascalVOC}}&DAG & FR  & 0.70   & 0.050   & 0.92 & ~~9.8  \\ 
        &UEA   & FR    & 0.70 & 0.050 & 0.93          & --     \\ 
        &SCA   & R18   & 0.67 & 0.060 & 0.91          & ~20.1  \\ 
        &SCA   & DLA34 & 0.77 & 0.110 & 0.86          & ~91.5  \\ 
        &DCA   & R18   & 0.67 & 0.070 & 0.90          & ~~~0.3 \\
        &DCA   & DLA34 & 0.77 & 0.050 & \textbf{0.94} & ~~~0.7 \\ \hline
     
        \multirow{4}{*}{\rotatebox[origin=c]{90}{MS-COCO}}
        &SCA       & R18   & 0.29  & 0.027 & 0.91          & ~50.4\\ 
        &SCA       & DLA34 & 0.37  & 0.030 & 0.92          & 216.0\\ 
        &DCA       & R18   & 0.29  & 0.002 & 0.99          & ~~~~\textbf{1.5}\\ 
        &DCA       & DLA34 & 0.37  & 0.002 & \textbf{0.99} & ~~~~2.4  \\ \hline
        \end{tabular}
    }
\vspace*{-0.5cm}
\end{table}

\noindent \textbf{White-Box Attack\footnote{More experimental results and hyperparameters analysis of DCA and SCA are included in the supplement material.}.} We have conducted white-box attacks on two popular object detection methods. Both use CenterNet but with different backbones: one, denoted as R18, with Resdcn18 \cite{he2016deep} and the other, DLA34 \cite{yu2018deep}, with Hourglass~\cite{newell2016stacked}. 

Table.~\ref{overall_white} shows the white-box attack results on both PascalVOC and MS-COCO. For comparison, it also contains the reported attack results of DAG and UEA attacking Faster-RCNN with VGG16~\cite{simonyan2014very} backbone, denoted as FR, on PascalVOC. There is no reported attack performance on MS-COCO for DAG and UEA.
UEA's average attack time in Table.~\ref{overall_white} is marked as ``--" (unavailable) 
because, as a GAN-based apporach, UEA's average attack time should include GAN's training time, which is unavailable. Compare with optimization-based attack methods~\cite{xie2017adversarial}, a GAN-based attack method consumes a lot of time for training and needs to retrain a new model to attack another task. Thus a GAN-based attack method sacrifices attack flexibility and cannot be used in some scenarios with high flexibility requirements.

The top half of Table.~\ref{overall_white} shows the attack performance on PascalVOC. We can see that: 
\textbf{(1)} DCA achieves higher ASR than DAG and UEA, and SCA achieves the best ASR performance.
\textbf{(2)} DCA is 14 times faster than DAG. We cannot compare with UEA since its attack time is unavailable. Qualitative comparison between DAG and our methods in shown in Fig.~\ref{fig: effi_framework}.
The bottom half of Table.~\ref{overall_white} shows the attack performance of our methods on MS-COCO.
SCA's ASR on both R18 and DLA34 is in the same ballpark as the ASR of DAG and UEA on PascalVOC, while DCA achieves the highest ASR, 99.0\%.
We conclude that both DCA and SCA achieve the state-of-the-art attack performance.

\begin{table}[t]
    \begin{center}
    \caption{\small
    Black-box attack results on the PascalVOC dataset. \textbf{From} in the leftmost column denotes the models where adversarial examples are generated from.  \textbf{To} in the top row means the attacked models that adversarial examples transfer to.}
    \label{tab_black_pascal}
    \resizebox{0.99\columnwidth}{!}{
        \begin{tabular}{l||c|c||c|c||c|c||c|c||c|c}
            \hline
            \multirow{2}{*}{\diagbox{\textbf{From}}{\textbf{To}}} & \multicolumn{2}{c||}{Resdcn18} & \multicolumn{2}{c||}{DLA34} & \multicolumn{2}{c||}{Resdcn101} & \multicolumn{2}{c||}{Faster-RCNN} & \multicolumn{2}{c}{SSD300} \\ \cline{2-11} 
                                   & mAP           & ATR       & mAP         & ATR      & mAP            & ATR           & mAP            & ATR           & mAP           & ATR           \\ \hline \hline
            Clean                     & 0.67          & --        & 0.77        & --       & 0.76           & --            & 0.71           & --            & 0.77          & --            \\ \hline
            DAG \cite{xie2017adversarial} & 0.65          & 0.19      & 0.75        & 0.16     & 0.74           & 0.16          & 0.60           & 1.00          & 0.76          & 0.08          \\ \hline
            R18-DCA                   & 0.10          & 1.00      & 0.62        & 0.23     & 0.65           & 0.17          & 0.61           & 0.17          & 0.72          & 0.08          \\ \hline
            DLA34-DCA                 & 0.50          & 0.28      & 0.07        & 1.00     & 0.62           & 0.2           & 0.53           & 0.28          & 0.67          & 0.14          \\ \hline
            R18-SCA                   & 0.31          & 1.00      & 0.62        & 0.36     & 0.61           & 0.37          & 0.55           & 0.42          & 0.70          & 0.17          \\ \hline
            DLA34-SCA                 & 0.42          & 0.90      & 0.41        & 1.00     & \textbf{0.53}  & \textbf{0.65} & \textbf{0.44}  & \textbf{0.82} & \textbf{0.62} & \textbf{0.42} \\ \hline
            \end{tabular}}
    \end{center}
\vspace{-0.75cm}
\end{table}

\begin{table}[t]
    \begin{center}
    \caption{\small Black-box attack results on the MS-COCO dataset. \textbf{From} in the leftmost column denotes the models where adversarial examples are generated from.  \textbf{To} in the top row means the attacked models that adversarial examples transfer to.}
    \label{tab_black_coco}
    \vspace{-0.25cm}
    \resizebox{\columnwidth}{!}{
        \begin{tabular}{l||c|c||c|c||c|c||c|c}
            \hline
            \multirow{2}{*}{\diagbox{\textbf{From}}{\textbf{To}}} & \multicolumn{2}{c||}{Resdcn18} & \multicolumn{2}{c||}{DLA34} & \multicolumn{2}{c||}{Resdcn101} & \multicolumn{2}{c}{CornerNet} \\ \cline{2-9} 
                                  & mAP           & ATR        & mAP         & ATR      & mAP            & ATR           & mAP            & ATR            \\ \hline \hline
            Clean                    & 0.29          & --         & 0.37        & --       & 0.37           & --            & 0.43           & --             \\ \hline
            R18-DCA                  & 0.01          & 1.00       & 0.29        & 0.21     & 0.28           & 0.25          & 0.38           & 0.12           \\ \hline
            DLA34-DCA                & 0.10          & 0.67       & 0.01        & 1.00     & 0.12           & 0.69          & 0.13           & 0.72           \\ \hline
            R18-SCA                  & 0.11          & 1.00       & 0.27        & 0.41     & 0.24           & 0.57          & 0.35           & 0.30           \\ \hline
            DLA34-SCA                & 0.07          & 0.92       & 0.06        & 1.00     & \textbf{0.09}  & \textbf{0.92} & \textbf{0.12}  & \textbf{0.88}  \\ \hline
        \end{tabular}}
    \end{center}
\vspace{-0.75cm}
\end{table}

\noindent \textbf{Black-Box Attack and Transferability.} 
Black-box attacks can be classified into two categories: cross-backbone and cross-network. 
For cross-backbone attacks, we evaluate the transferability with Resdcn101~\cite{he2016deep} on PascalVOC and MS-COCO. For cross-network attack, we evaluate with not only anchor-free object detector CornerNet~\cite{law2019cornernet} but also two-stage anchor-based detectors, Faster-RCNN~\cite{ren2015faster} and SSD300~\cite{liu2016ssd}. Faster-RCNN and SSD300 are tested on PascalVOC. CornerNet is tested on MS-COCO with backbone Hourglass~\cite{newell2016stacked}. 

To simulate a real-world attack transferring scenario, we generate adversarial examples on the CenterNet and save them in the JPEG format, which may cause them to lose the ability to attack target models~\cite{dziugaite2016study} as some key detailed information may get lost due to the lossy JPEG compression. Then, we reload them to attack target models and compute $mAP$. This process has a more strict demand on adversarial examples but should improve their transferability.

i) \noindent \textit{Attack transferability on PascalVOC.} Adversarial examples are generated on CenterNet with Resdcn18 and DLA34 backbones for both SCA and DCA. For comparison, DAG is also used to generate adversarial examples on Faster-RCNN. 
These adversarial examples are then used to attack the other four models. All the five models are trained on PascalVOC. Table.~\ref{tab_black_pascal} shows the experimental results. We can see from the table that adversarial examples generated by our method can successfully transfer to not only CenterNet with different backbones but also completely different types of object detectors, Faster-RCNN and SSD. We can also see that DCA is more robust to the JPEG compression than SCA, while SCA achieves higher ATR than DCA in the black-box test. Table.~\ref{tab_black_pascal} indicates that DAG is sensitive to the JPEG compression, especially when its adversarial examples are used to attack Faster-RCNN, and has a very poor transferability in attacking CenterNet and SSD300. We conclude that both DCA and SCA perform better than DAG on both transferability and robustness to the JPEG compression.

ii) \noindent \textit{Attack Transferability on MS-COCO.} Similar to the above experiments, adversarial examples are generated on Centernet with Resdcn18 and DLA34 backbones and then used to attack other object detection models. The experimental results are summarized in Table.~\ref{tab_black_coco}. The table indicates that generated adversarial examples can attack not only  CenterNet with different backbones but also CornerNet.

\begin{table}[t]\small
    \centering
    \caption{\small Perceptibility of the perturbation.}
    \label{qa}
    \resizebox{0.75\columnwidth}{!}{
        \begin{tabular}{l||l|l}
        \hline
        Network & $\mathop{P_{L_2}}$  & $\mathop{P_{L_0}}$ \\ \hline \hline
        DAG & $\mathop{2.8 \times 10^{-3} }$ & $\mathop{\geq\ 99.0\%}$ \\
        R18-Pascal & $\mathop{5.1\times 10^{-3}}{(DCA)}$ & $\mathop{0.22\%} {(SCA)}$ \\
        DLA34-Pascal    & $\mathop{5.1\times 10^{-3}}{(DCA)}$ & $\mathop{0.27\%}{(SCA)}$ \\
        R18-COCO   & $\mathop{4.8\times 10^{-3}}{(DCA)}$ & $\mathop{0.39\%}{(SCA)}$ \\
        DLA34-COCO      & $\mathop{5.2\times 10^{-3}}{(DCA)}$ & $\mathop{0.65\%}{(SCA)}$ \\ \hline
        \end{tabular}}
\vspace{-0.4cm}
\end{table}

\noindent \textbf{Perceptibility.}
\label{sec: perceptibility}
The perceptibility results of adversarial perturbations of DCA and SCA are shown on Table.~\ref{qa}.
We can see that $\mathop{P_{L_0}}$ of SCA is lower than 1\%, meaning that SCA can fool the detectors by perturbing only a few number of pixels.
Although DCA has a higher $\mathop{P_{L_2}}$ than DAG, perturbations generated by DCA are still hard for humans to perceive.
We also provide qualitative examples for comparison in Fig.~\ref{fig: effi_framework}.

\section{Conclusion}
\label{sec: conclusion}
\noindent

In this paper, we propose a category-wise attack to attack anchor-free object detectors. To the best of our knowledge, it is the first adversarial attack on anchor-free object detectors. Our attack manifests in two forms, SCA and DCA, when minimizing the $L_0$ and $L_\infty$ norms, respectively. Both SCA and DCA focus on global and high-level semantic information to generate adversarial perturbations. Our experiments with CenterNet on two public object detection benchmarks indicate that both SCA and DCA achieve the state-of-the-art attack performance and transferability.


\end{document}